# Automatic segmentation of the spinal cord and intramedullary multiple sclerosis lesions with convolutional neural networks

*September 11th, 2018*
*Under review - NeuroImage*

## Authors:

Charley Gros [1], Benjamin De Leener [1], Atef Badji [1,2], Josefina Maranzano [4], Dominique Eden [1], Sara M. Dupont [1,3], Jason Talbott [3], Ren Zhuoquiong [5], Yaou Liu [5,6], Tobias Granberg [7,8], Russell Ouellette [7,8], Yasuhiko Tachibana [22], Masaaki Hori [23], Kouhei Kamiya [23], Lydia Chougar [23,24], Leszek Stawiarz [7], Jan Hillert [7], Elise Bannier [9,10], Anne Kerbrat [10,11], Gilles Edan [10,11], Pierre Labauge [12], Virginie Callot [13,14], Jean Pelletier [14,15], Bertrand Audoin [14,15], Henitsoa Rasoanandrianina [13,14], Jean-Christophe Brisset [16], Paola Valsasina [17], Maria A. Rocca [17], Massimo Filippi [17], Rohit Bakshi [18], Shahamat Tauhid [18], Ferran Prados [19,26], Marios Yiannakas [19], Hugh Kearney [19], Olga Ciccarelli [19], Seth Smith [20], Constantina Andrada Treaba [8], Caterina Mainero [8], Jennifer Lefeuvre [21], Daniel S. Reich [21], Govind Nair [21], Vincent Auclair [27], Donald G. McLaren [27], Allan R. Martin [28], Michael G. Fehlings [28], Shahabeddin Vahdat [29,25], Ali Khatibi [4,25], Julien Doyon [4,25], Timothy Shepherd [30], Erik Charlson [30], Sridar Narayanan [4], Julien Cohen-Adad [1,25]

## Affiliations:

[1] NeuroPoly Lab, Institute of Biomedical Engineering, Polytechnique Montreal, Montreal, QC, Canada
[2] Department of Neuroscience, Faculty of Medicine, University of Montreal, Montreal, QC, Canada
[3] Department of Radiology and Biomedical Imaging, Zuckerberg San Francisco General Hospital, University of California, San Francisco, CA, USA
[4] McConnell Brain Imaging Centre, Montreal Neurological Institute, Montreal, Canada
[5] Department of Radiology, Xuanwu Hospital, Capital Medical University, Beijing 100053, P. R. China
[6] Department of Radiology, Beijing Tiantan Hospital, Capital Medical University, Beijing 100050, P. R. China
[7] Department of Clinical Neuroscience, Karolinska Institutet, Stockholm, Sweden
[8] Martinos Center for Biomedical Imaging, Massachusetts General Hospital, Boston, USA
[9] CHU Rennes, Radiology Department
[10] Univ Rennes, Inria, CNRS, Inserm, IRISA UMR 6074, Visages U1128, France
[11] CHU Rennes, Neurology Department
[12] MS Unit. DPT of Neurology. University Hospital of Montpellier

[13] Aix Marseille Univ, CNRS, CRMBM, Marseille, France
[14] APHM, CHU Timone, CEMEREM, Marseille, France
[15] APHM, Department of Neurology, CHU Timone, APHM, Marseille
[16] Observatoire Français de la Sclérose en Plaques (OFSEP) ; Univ Lyon, Université Claude Bernard Lyon 1 ; Hospices Civils de Lyon ; CREATIS-LRMN, UMR 5220 CNRS & U 1044 INSERM ; Lyon, France
[17] Neuroimaging Research Unit, INSPE, Division of Neuroscience, San Raffaele Scientific Institute, Vita-Salute San Raffaele University, Milan, Italy
[18] Brigham and Women's Hospital, Harvard Medical School, Boston, USA
[19] Queen Square MS Centre, UCL Institute of Neurology, Faculty of Brain Sciences, University College London, London (UK)
[20] Vanderbilt University, Tennessee, USA
[21] National Institute of Neurological Disorders and Stroke, National Institutes of Health, Maryland, USA
[22] National Institute of Radiological Sciences, Chiba, Chiba, Japan
[23] Juntendo University Hospital, Tokyo, Japan
[24] Hospital Cochin, Paris, France
[25] Functional Neuroimaging Unit, CRIUGM, Université de Montréal, Montreal, QC, Canada
[26] Center for Medical Image Computing (CMIC), Department of Medical Physics and Biomedical Engineering, University College London, London, United Kingdom
[27] Biospective Inc., Montreal, QC, Canada
[28] Division of Neurosurgery, Department of Surgery, University of Toronto, Toronto, ON, Canada
[29] Neurology Department, Stanford University, US
[30] NYU Langone Medical Center, New York, USA

# Corresponding author:

Julien Cohen-Adad
Dept. Genie Electrique, L5610
Ecole Polytechnique
2900 Edouard-Montpetit Bld
Montreal, QC, H3T 1J4, Canada
Phone: 514 340 5121 (office: 2264); Skype: jcohenadad; e-mail: jcohen@polymtl.ca

# Abstract

The spinal cord is frequently affected by atrophy and/or lesions in multiple sclerosis (MS) patients. Segmentation of the spinal cord and lesions from MRI data provides measures of damage, which are key criteria for the diagnosis, prognosis, and longitudinal monitoring in MS. Automating this operation eliminates inter-rater variability and increases the efficiency of large-throughput analysis pipelines. Robust and reliable segmentation across multi-site spinal cord data is challenging because of the large variability related to acquisition parameters and image artifacts. In particular, a precise delineation of lesions is hindered by a broad heterogeneity of lesion contrast, size, location, and shape. The goal of this study was to develop a fully-automatic framework — robust to variability in both image parameters and clinical condition — for segmentation of the spinal cord and intramedullary MS lesions from conventional MRI data of MS and non-MS cases. Scans of 1,042 subjects (459 healthy controls, 471 MS patients, and 112 with other spinal pathologies) were included in this multi-site study (n=30). Data spanned three contrasts ($T_1$-, $T_2$-, and $T_2^*$-weighted) for a total of 1,943 volumes and featured large heterogeneity in terms of resolution, orientation, coverage, and clinical conditions. The proposed cord and lesion automatic segmentation approach is based on a sequence of two Convolutional Neural Networks (CNNs). To deal with the very small proportion of spinal cord and/or lesion voxels compared to the rest of the volume, a first CNN with 2D dilated convolutions detects the spinal cord centerline, followed by a second CNN with 3D convolutions that segments the spinal cord and/or lesions. CNNs were trained independently with the Dice loss. When compared against manual segmentation, our CNN-based approach showed a median Dice of 95% vs. 88% for *PropSeg* ($p \leq 0.05$), a state-of-the-art spinal cord segmentation method. Regarding lesion segmentation on MS data, our framework provided a Dice of 60%, a relative volume difference of -15%, and a lesion-wise detection sensitivity and precision of 83% and 77%, respectively. In this study, we introduce a robust method to segment the spinal cord and intramedullary MS lesions on a variety of MRI contrasts. The proposed framework is open-source and readily available in the Spinal Cord Toolbox.

## Abbreviations:

**CNN**: convolutional neural network ; **IQR**: interquartile range ; **MS**: multiple sclerosis ; **MSE**: mean square error ; **SCT**: spinal cord toolbox ; **SVM**: support vector machine.

# 1. Introduction

Multiple sclerosis (MS) is a chronic immune mediated disease of the central nervous system, with variable clinical expression. The pathologic hallmark of MS is the occurrence of focal areas of inflammatory demyelination within the brain and spinal cord, known as lesions (Popescu and Lucchinetti, 2012). MS lesions exhibit variable degrees of demyelination, axonal injury and loss, remyelination, and gliosis. Impaired axonal conduction often causes motor, sensory, visual, and cognitive impairment (Compston and Coles, 2002). Clinicians and researchers extensively use conventional MRI (e.g., T2-weighted) to non-invasively quantify the lesion burden in time and space (Filippi and Rocca, 2007; Kearney et al., 2015b; Simon et al., 2006; Sombekke et al., 2013; Weier et al., 2012). The study of spinal cord lesions has recently garnered interest (Hua et al., 2015; Kearney et al., 2015a) given its potential value for diagnosis and prognosis of MS (Arrambide et al., 2018; Sombekke et al., 2013; Thorpe et al., 1996). Moreover, spinal cord atrophy is common in MS (Bakshi et al., 2005), and the quantification of such atrophy is clinically relevant and correlates with clinical disability (Cohen et al., 2012; Kearney et al., 2014; Losseff et al., 1996; Lundell et al., 2017; Rocca et al., 2013, 2011). Consequently, segmentation of the spinal cord and MS lesions contained within it (intramedullary lesions) is a common procedure to quantitatively assess the structural integrity of this portion of the central nervous system in MS patients. However, manual segmentation is time-consuming and suffers from intra- and inter-rater variability. Hence, there is a need for robust and automatic segmentation tools for the spinal cord and the intramedullary MS lesions.

Various automatic spinal cord segmentation methods have been proposed in the past few years, including active contours and surface-based approaches (De Leener et al., 2015; Koh et al., 2010), and atlas-based methods (Carbonell-Caballero et al., 2006; Chen et al., 2013; Pezold et al., 2015; Tang et al., 2013). While these methods have shown good performance (De Leener et al., 2016), they often require a specific region of interest and/or are limited to a specific contrast and resolution. Moreover, the lack of validation against multi-site data or cases with spinal cord damage has limited their application in large clinical multi-site studies. Automatic spinal cord segmentation is difficult to achieve robustly and accurately across the broad range of spinal cord shapes, lengths, and pathologies; and across variable image dimensions, resolutions, orientations, contrasts, and artifacts (e.g. susceptibility, motion, chemical shift, ghosting, blurring, Gibbs). Figure 1 illustrates these challenges, depicting the heterogeneity frequently observed in multi-site clinical spinal cord data sets.

The automatic segmentation of MS lesions has been thoroughly investigated over the past two decades for brain data sets (García-Lorenzo et al., 2013; Lladó et al., 2012), although it still remains a challenging task (Meier et al., 2018; Roy et al., 2018; Valverde et al., 2017a, 2017b). While previous methods have shown reasonable performance in the brain, they are not easily transposable to the spinal cord, mainly because of its specific morphology. Furthermore, traditional intensity-based segmentation methods are challenging in spinal cord images because of (i) the frequent intensity bias field in the Superior-to-Inferior axis which is difficult to correct, (ii) the confounding of lesion intensities with those of normal structures (e.g. grey matter on $T_2$*-weighted images), or artifacts, and (iii) partial volume effects, where several structures may contribute to the signal of border voxels (e.g. cerebrospinal fluid and cord). To provide an overview of these challenges, Figure 1 shows instances of intramedullary MS lesions exhibiting heterogeneity (i.e. location, size, and shape), along with their intensity histograms which demonstrate a large overlap with the spinal cord intensities.

The last years have witnessed a noteworthy interest in convolutional neural networks (CNNs) for image segmentation tasks, with remarkable performance in different domains, notably in medical image analysis (Litjens et al., 2017). The game-changing advantage of CNNs, compared to feature engineering based approaches, is their hierarchical representation learning strategy to find appropriate filters on their own. Indeed, the features learned in the first layers come together and make abstract shapes, which often have meaning in their deeper layers. CNN methods have proven to be highly robust to varying image appearances. In particular, since 2015, U-net architecture achieved a notable breakthrough in the biomedical image segmentation community (Ronneberger et al., 2015), even for tasks with little available annotated training data. The good performance of the U-net architecture is often explained by the use of two distinct paths: a contracting path to capture context, followed by a symmetric expanding path to recover the spatial information, with the support of skip connections between the paths. However, training CNNs on very unbalanced data sets, such as those encountered in MS spinal cord lesion segmentation tasks (i.e. data with < 1% of lesion voxels), remains a focus of active research (Buda et al., 2017; Sudre et al., 2017).

In this work, we propose an original and fully automatic framework for segmenting the spinal cord and/or intramedullary MS lesions from a variety of MRI contrasts and resolutions. The presented methods are based on a sequence of CNNs, specifically designed for spinal cord morphometry. We trained the networks and evaluated the robustness of the framework using a multi-site clinical data set ($n_{volumes}$=1,943), which features a variety of pathologies, artifacts, contrasts, resolutions, dimensions, and orientations.

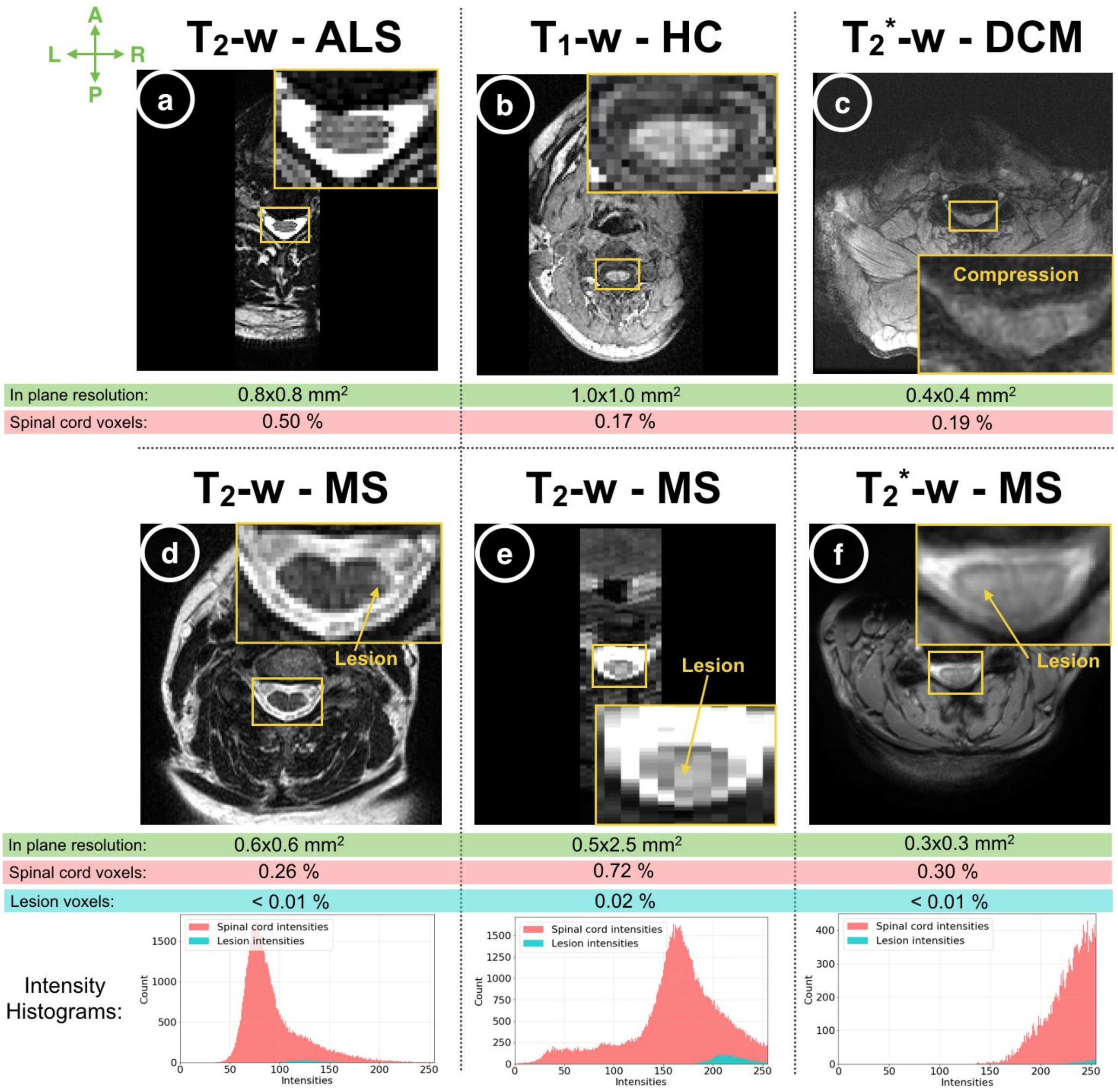

**Figure 1**: **Spinal cord axial slice samples.** (a-f) show the variability of the images in terms of resolution, field of view, and MR contrasts. Images were acquired from 6 different sites, of subjects with different clinical status: healthy control (HC, b), amyotrophic lateral sclerosis (ALS, a), degenerative cervical myelopathy (DCM, c) and multiple sclerosis (MS, d-f). The in-plane resolutions vary across the images. For all images, the spinal cord and lesion voxels represent less than 1% and 0.1%, respectively, of the entire volume. The shape, location, size, and level of contrast differ among MS lesions (d-f). The histograms for spinal cord and lesion voxels of the MS patient (d-f) images are shown at the bottom. Although lesions mostly appear hyperintense in $T_2$- and $T_2^*$-weighted, a substantial overlap between spinal cord and lesion intensities is observed, leading to low contrast, especially for $T_2^*$-w images (f) with similarities between grey matter and lesion appearance.

# 2. Materials and Methods

## 2.1. Data

Thirty centers contributed to this study, gathering retrospective 'real world' data from 1,042 subjects, including healthy controls (n=459), patients with MS or suspected MS (n=471), as well as degenerative cervical myelopathy (n=55), neuromyelitis optica (n=19), spinal cord injury (n=4), amyotrophic lateral sclerosis (n=32), and syringomyelia (n=2). The MS cohort spanned a large heterogeneity of clinical conditions in terms of the Expanded Disability Status Scale (mean: 2.5 ; range: 0-8.5) and phenotype: clinically isolated syndrome (n=29), relapsing-remitting MS (n=283), secondary progressive MS (n=76), and primary progressive MS (n=69). Clinical data were not available for all MS patients. Images were acquired at 3T and 7T on various platforms (Siemens, Philips and GE). Contrasts included $T_2$−weighted ($n_{vol.}$ = 904), $T_1$−weighted ($n_{vol.}$ = 151), and $T_2$*-weighted ($n_{vol.}$ = 888). The coverage substantially differed among subjects, with volumes including the brain and/or diverse vertebral levels (cervical, thoracic, lumbar). Spatial resolutions included isotropic ($n_{vol.}$ = 451, from 0.7 to 1.3mm) and anisotropic data with axial ($n_{vol.}$ = 1010, in plane: from 0.2 to 0.9mm, slice thickness including slice gap: from 1.0 to 24.5mm), or with sagittal orientation ($n_{vol.}$ = 482, in plane: from 0.4 to 1.1mm, slice thickness: from 0.8 to 5.2mm). Figure 2 summarises the data set, while Table A1 (see Appendix) details the imaging parameters across participating sites.

Four trained raters (BDL, SD, DE, CG) manually corrected the segmentation produced by *PropSeg* (De Leener et al., 2014) using FSLview (Jenkinson et al., 2012). The resulting spinal cord mask was considered as ground-truth and is herein referred to as "manual segmentation". Using data from MS patients ($n_{vol.}$=967), lesion masks were generated by 7 raters including radiologists (JM, JT, MH, YT, RZ, LC) and trained (AB) raters using ITK-SNAP Toolbox 3.6.0 (Yushkevich and Gerig, 2017). Image raters were blind to diagnostic and clinical information. Guidelines followed by raters are available at: osf.io/d4evy/. Among the MS volumes segmented by the raters, 17.7% ($n_{vol.}$=171) were considered lesion free. The lesion involvement was highly heterogeneous across patients, with a mean (range) lesion count of 3.1 (0-17) and total lesion volume of 192mm$^3$ (0.0-1679.8mm$^3$). Over the entire MS data set, 0.01% of image voxels on average were confirmed to contain lesions by the experts, showing the unbalanced nature of the data.

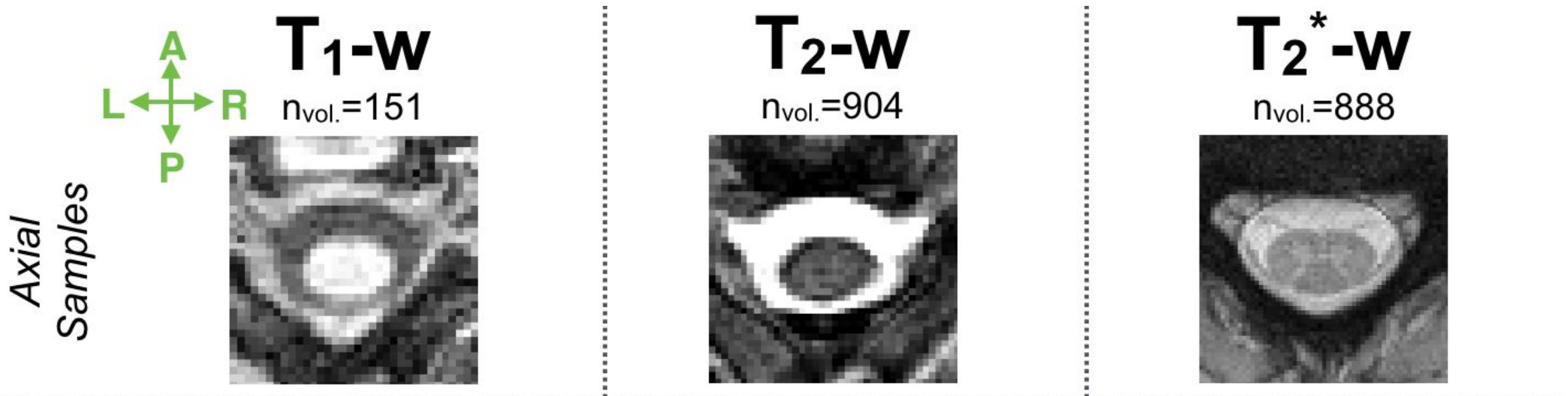

| | T1-w Isotropic | T1-w Anisotropic Ax. Orient. | T1-w Anisotropic Sag. Orient. | T2-w Isotropic | T2-w Anisotropic Ax. Orient. | T2-w Anisotropic Sag. Orient. | T2*-w Isotropic | T2*-w Anisotropic Ax. Orient. | T2*-w Anisotropic Sag. Orient. |
|---|---|---|---|---|---|---|---|---|---|
| n | 149 | 1 | 1 | 302 | 121 | 481 | 0 | 888 | 0 |
| In Plane Resolution | 1.0 [0.8, 1.3] | 0.4 [0.4, 0.4] | 0.4 [0.4, 0.4] | 0.9 [0.7, 1.0] | 0.5 [0.3, 0.9] | 0.6 [0.4, 1.1] | - | 0.4 [0.2, 0.9] | - |
| Slice Thickness | 1.0 [0.8, 1.3] | 1.0 [1.0, 1.0] | 0.9 [0.9, 0.9] | 0.9 [0.7, 1.0] | 3.3 [3.0, 7.5] | 2.9 [0.8, 5.2] | - | 4.7 [1.0, 24.5] | - |

Data indicate mean and range values in mm.

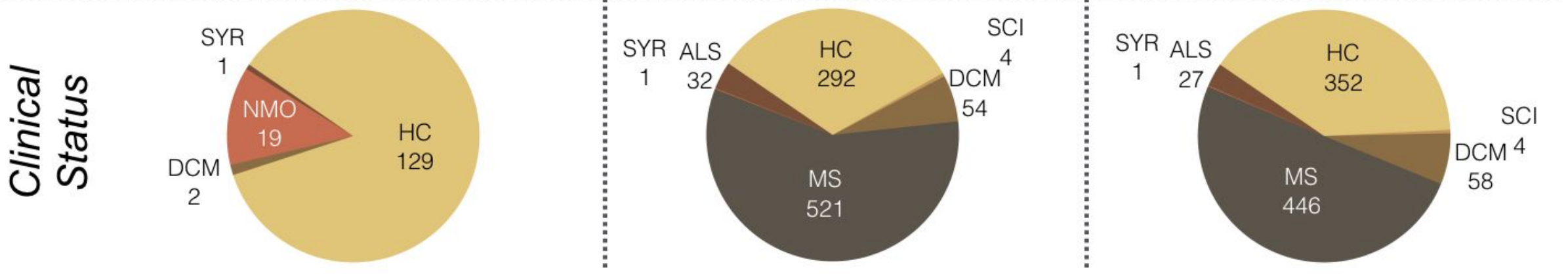

**Figure 2: Overview of the data set.** Samples of cross-sectional axial slices of the three MR contrast data sets ($T_1$-weighted, $T_2$-weighted, $T_2^*$-weighted) are depicted (top row). Image characteristics in terms of orientation (orient.) and resolution (resol.), grouped by isotropic, anisotropic and with axial (Ax.) orientation or sagittal (Sag.) orientation are presented (middle row). The last row shows the proportion of clinical status among the imaged subjects, including: healthy controls (HC), multiple sclerosis (MS), degenerative cervical myelopathy (DCM), neuromyelitis optica (NMO), traumatic spinal cord injury (SCI), amyotrophic lateral sclerosis (ALS), and syringomyelia (SYR). Imaging parameters across participating sites are detailed in Table A1 (see Appendix).

## 2.2. Segmentation framework

The proposed segmentation framework is depicted in Figure 3. The workflow consists of two major stages. The first stage detects the spinal cord centerline (Figure 3, step 1-2) and the second stage performs the spinal cord and/or lesion segmentation along the centerline (Figure 3, steps 3).

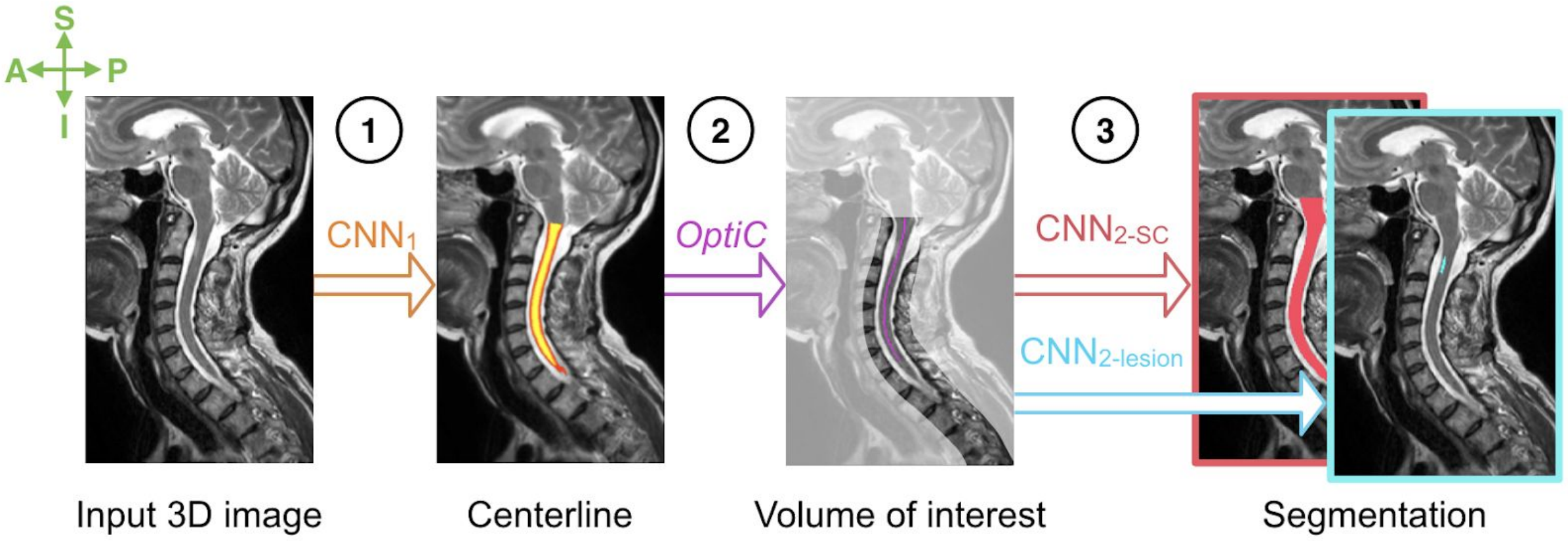

**Figure 3: Automatic segmentation framework.** (1) detection of the spinal cord by $CNN_1$ which outputs a heatmap (red-to-yellow) of the spinal cord location, (2) computation of the spinal cord centerline (pink) from the spinal cord heatmap (Gros et al., 2018), and extraction of 3D patches in a volume of interest surrounding the spinal cord centerline, (3) segmentation of the spinal cord (red) by $CNN_{2\text{-}SC}$, and/or of lesions (blue) by $CNN_{2\text{-}lesion}$. SC: Spinal cord ; CNN: Convolutional Neural Network ; S: Superior ; I: Inferior ; A: Anterior ; P: Posterior.

### 2.2.1. Sequential framework

CNNs can easily overfit because of two main features of our data set: (i) the high class imbalance due to the small number of voxels labeled as positive (~ 0.34% for spinal cord, ~ 0.01% for lesions), and (ii) the limited number of available labeled images. To prevent overfitting, the proposed framework split the learning scheme into two stages, each containing a CNN. The first stage consists of detecting the center of the spinal cord ($CNN_1$) and crop the image around it, while the second stage segments the spinal cord ($CNN_{2\text{-}SC}$) and/or the MS lesion ($CNN_{2\text{-}lesion}$). Note that $CNN_{2\text{-}SC}$ and $CNN_{2\text{-}lesion}$ were independently trained and can be run separately. The motivation behind the sequential approach is that CNNs have been shown to learn a hierarchical representation of the provided data since the stacked layers of convolutional filters are tailored towards the desired segmentation (Christ et al., 2017; LeCun et al., 2015; Valverde et al., 2017a). The designed sequential framework ensures that (i) $CNN_1$ learns filters to discriminate between the axial patches that contain spinal cord voxels versus patches that do not, (ii) while $CNN_{2\text{-}SC}$ (and $CNN_{2\text{-}lesion}$ ) is trained to optimise a set of filters tailored to the spinal cord (and the lesions) from training patches centered around the spinal cord.

Automatic preprocessing steps include resampling to 0.5mm isotropic images (based on preliminary optimisations), and matrix re-orientation (RPI, i.e. Right-to-left, Posterior-to-anterior, Inferior-to-superior).

### 2.2.2. Spinal cord centerline detection

Detection of the cord centerline (Figure 3, step 1) is achieved with a 2D CNN ($CNN_1$), through each cross-sectional slice of the input volume.

For each input volume, we extract 2D patches (96x96) from the cross-sectional slices. We computed the mean intensity and standard deviation across the training patches, to normalise all the processed patches (i.e. zero mean and unit variance), including the validation and testing patches.

$CNN_1$ architecture was adapted from the U-net architecture (Ronneberger et al., 2015) by reducing the downsampling layers from four to two layers, and by replacing conventional convolutions with dilated convolutions in the contracting path. Briefly, dilated convolution is a convolution with defined gaps, which provides an exponential expansion of the receptive view with a linear increase of parameters (Yu and Koltun, 2015). The motivation behind the use of dilated convolutions is to capture more contextual information (i.e. broader view of the input), with fewer parameters compared to a conventional solution, which involves additional downsampling layers. Preliminary experiments led us to use a dilation rate of three (i.e. a gap of two pixels per input, as also illustrated in Figure 1 of (Yu and Koltun, 2015)). To reduce overfitting, Batch Normalisation (Ioffe and Szegedy, 2015), rectified linear activation function (Nair and Hinton, 2010), and Dropout (training with p=0.2) (Srivastava et al., 2014) follow each convolution layer.

Training of $CNN_1$ was performed on each contrast data set separately (i.e. three trained models: $T_1$-w, $T_2$-w, and $T_2^*$-w), using the Adam optimizer (Kingma and Ba, 2014), with a learning rate of 0.0001, a batch size of 32, and 100 epochs. We employed Dice loss (Milletari et al., 2016) for the loss function due to its insensitivity to high class imbalance, as favoured by recent studies dealing with this issue (Drozdzal et al., 2018; Perone et al., 2017; Sudre et al., 2017). We performed an extensive data augmentation of the training samples, including shifting (±10 voxels in each direction), flipping, rotation (±20° in each direction), and elastic deformations (Simard et al., 2003) (deformation coefficient of 100, standard deviation of 16). Elastic transformations were shown to be efficient at increasing learning invariance (Dosovitskiy et al., 2014) and realistic variation in tissue (Ronneberger et al., 2015).

Spinal cord centerline extraction is achieved by reconstructing a volume from the patch inference of $CNN_1$, where values indicate the degree of confidence regarding the spinal cord location. Because $CNN_1$ outputs a prediction mask with abrupt boundaries, we compute the Euclidean distance map from the $CNN_1$ output to assist with spinal cord centerline detection (red-to-yellow values in Figure 3, step 1). We infer the centerline from this spinal cord distance map using *OptiC* (Gros et al., 2018), a previously published fast global-curve optimisation algorithm, which regularises the centerline continuity along the Superior-to-Inferior axis (pink centerline in Figure 3, step 2).

### 2.2.3. Spinal cord and MS lesions segmentation

Segmentation of the spinal cord and the intramedullary lesions are achieved by $CNN_{2\text{-}SC}$ and $CNN_{2\text{-}lesion}$, which both are 3D CNNs investigating in a volume of interest surrounding the inferred cord centerline.

From each volume, we extract 3D patches along the spinal cord centerline (Figure 3, step 2) with the following sizes: 64x64x48 for the spinal cord (i.e. $CNN_{2\text{-}SC}$) and 48x48x48 for MS lesions (i.e. $CNN_{2\text{-}Lesion}$). In preliminary experiments, we investigated different patch sizes (32x32x32, 48x48x48, 64x64x48, and 96x96x48) and decided on a compromise between the class imbalance, the risk of overfitting, and the computational cost. We apply an intensity normalisation algorithm on the stacked patches of each volume to homogenise the intensity distributions on a standardised intensity range (Nyúl and Udupa, 1999; Pereira et al., 2016; Shah et al., 2011). Finally, following the same process as in section 2.2.2, we normalise the patch intensities by centering the mean and normalising the standard deviation.

$CNN_{2\text{-}SC}$ and $CNN_{2\text{-}Lesion}$ architectures draw from the 3D U-net scheme (Çiçek et al., 2016); however, we reduced the depth of the U-shape from three to two, thus limiting the number of parameters and the amount of memory required for training.

Training of $CNN_{2\text{-}SC}$ and $CNN_{2\text{-}lesion}$ were also undertaken for each contrast, even though $CNN_{2\text{-}lesion}$ was trained with MS data only. We trained the models using the Adam optimizer, the Dice loss, the Dropout (p=0.4), and the following parameters: a batch size of 4, learning rate of $5x10^{-5}$, and total number of epochs of 300. Besides flipping operations, the data augmentation procedure included small local erosions and dilations of the manual lesion edges, which serve to test the confidence of the network on subjective lesion borders.

During the inference stage, $CNN_{2\text{-}SC}$ and $CNN_{2\text{-}Lesion}$ independently segment 3D patches extracted from a testing data. We apply a threshold of 0.5 to the CNNs predictions before reconstructing a 3D volume (Figure 3, step 4). The presented framework does not contain additional post-processing.

## 2.3. Implementation

We implemented the proposed method in the Python 2.7 language, using Keras[1] (v2.6.0) and TensorFlow[2] (v1.3.0) libraries. The code of the CNNs implementations is available at [URL][3]. Moreover, the presented methods are readily available through the functions sct_deepseg_sc and sct_deepseg_lesion as part of the Spinal Cord Toolbox (SCT) (De Leener et al., 2017a) version v3.2.2 and higher. These functions are robust to any image resolution and orientation, as well as number of slices, even for single axial slice images.

CNN training was carried out on a single NVIDIA Tesla P100 GPU with 16GB RAM memory and took approximately 6, 70, and 102 hours, for $CNN_1$, $CNN_{2\text{-}SC}$, and $CNN_{2\text{-}lesion}$, respectively. Training was stopped when the training loss kept decreasing while the validation loss steadily increased or settled down. Contrary to the training which requires high computational power such as that offered by a GPU, inference (i.e. segmentation) can run in only a few minutes on a standard CPU.

[1] https://keras.io/, version 2.6.0
[2] https://www.tensorflow.org/, version 1.3.0
[3] https://github.com/neuropoly/spinalcordtoolbox/tree/master/spinalcordtoolbox/deepseg_sc

## 2.4. Evaluation

For each contrast (i.e. $T_1$-, $T_2$-, $T_2^*$-weighted), the networks were trained on 80% of the subjects, with 10% held out for validation and 10% for testing (i.e. for results presented in section 3.). In particular, the testing data set contained data from two sites (n=57), which were not present during the training procedure, in order to evaluate the generalisation of the pipeline to new image features.

### 2.4.1. Spinal cord centerline detection

We evaluated the cord centerline detection (i.e. output of *OptiC,* see Figure 3, step 1-2), by computing (i) the Mean Square Error (MSE) between the predicted and manual spinal cord centerlines, (ii) the localization rate, defined as the percentage of axial slices for which the predicted centerline was included in the manually-segmented spinal cord. We generated the manual spinal cord centerlines by computing the center of mass of each axial slice of the manual spinal cord segmentations, regularised with an approximated non-uniform rational bezier spline, as described in (De Leener et al., 2017b).

We compared our spinal cord detection method (Figure 3, step 1-2) to a recently-published study (Gros et al., 2018) that introduced a global curve optimisation algorithm (*OptiC*, Figure 3, Step 2) but used a trained Support-Vector-Machine (SVM) algorithm to produce the spinal cord heatmap (instead of the $CNN_1$ at Step 1). We refer to this as "*SVM+OptiC*" in the remainder of this work. A non-parametric test (Kruskal-Wallis) was applied to assess potential performance differences between these two approaches.

### 2.4.2. Spinal cord segmentation

We assessed the spinal cord segmentation performance (i.e. output of $CNN_{2\text{-}SC}$, see Figure 3, step 3), by calculating (i) the Dice Similarity Coefficient (Dice, 1945) and (ii) the relative volume difference in segmented volume (asymmetric metric) between the automatic and the manual segmentation masks. We compared the spinal cord segmentation method to a previously-published unsupervised method, *"PropSeg"*, which is based on multi-resolution propagation of tubular deformable models (De Leener et al., 2015). Kruskal-Wallis tests assessed performance differences between the two methods.

### 2.4.3. MS lesion segmentation

We estimated the intramedullary MS lesion segmentation performance (i.e. output of $CNN_{2\text{-}lesion}$, see Figure 3, step 3), by calculating (i) the Dice, (ii) the relative volume difference, (iii) the voxel-wise sensitivity, and (iv) the voxel-wise precision between the automatic and the manual segmentation masks of the MS cohort. Voxel-wise metrics considered a voxel as correctly segmented by the algorithm (i.e. true positive) if it was labelled as "lesion" by the raters.

We also computed the lesion-wise sensitivity and the lesion-wise precision, where individual lesions (i.e. 3D connected objects) were analysed as entities (i.e. instead of each voxel separately, as for the voxel-wise metrics). We considered a candidate lesion as correctly detected (i.e. true positive) when the automatic segmentation connected-voxels overlapped with more than 25% of the manual segmentation voxels, otherwise it was considered as incorrectly detected (i.e. false positive). If a confirmed lesion (i.e. manually labelled) had an insufficient overlap (<25%) with the automatic segmentation voxels, then we defined it as not-detected (i.e. false negative).

The specificity of the automatic lesion detector was computed on data from healthy controls and MS patients who did not have any intramedullary lesion detected, and called volume-wise specificity in the remaining of this paper. We considered a volume as incorrectly detected (i.e. false positive) if at least one lesion was automatically detected. We assumed healthy control data to be lesion free.

#### 2.4.4. Inter-rater variability of the MS lesion segmentation

We estimated the inter-rater variability of lesion segmentation among all participating raters (n=7), on a randomised subset of patients (n=10). For each of these patients, two scans were available, which allows the raters to segment both scans in parallel by combining their information. For this purpose, we calculated the Dice coefficient between each rater's segmentation and a consensus reading mask, produced using "majority voting" across all the raters' labels.

# 3. Results

## 3.1. Spinal cord centerline detection

Table 1 (A.) presents the medians and interquartile ranges (IQRs) of the metrics evaluating the spinal cord centerline detection across contrasts. When averaging the performance metrics across all contrasts, the centerline detection using $CNN_1$ significantly outperformed the SVM-based method (p<0.001), as shown by the median MSE (IQR) of 1.0 (0.8) mm versus 5.5 (9.7) mm. While the two approaches produced similar results on 3D isotropic resolution and axial scans, CNN performed better on sagittal scans: median MSE 1.1 (0.9) mm for "*$CNN_1$+OptiC*" versus 11.6 (11.4) mm for "*SVM+OptiC*" (p<0.001). In volumes that included part of the brain, the method accurately confined the segmentation to between the top of C1 and pontomedullary junction (i.e. differentiated brain and spinal regions) in 87.0% of cases. The median MSE was largely improved by resorting to the curve optimisation algorithm, especially on degenerative cervical myelopathy patients, as it considerably decreased from 24.04mm ($CNN_1$ output, Figure 3, step 1) to 1.14mm ("*$CNN_1$+OptiC*" output, Figure 3, step 2).

## 3.2. Spinal cord segmentation

Figure 4 illustrates qualitative samples of spinal cord segmentation from the testing data set, comparing the manual against the automatic delineation. From visual inspection, the proposed method achieved encouraging results on (i) compressed and atrophied cords (e.g., see *S5_DCM17, S5_DCM2, S25_ALS5*), (ii) slices with poor contrast between cord and surrounding structures like cerebrospinal fluid (*S16_HC1)* or MS lesions (*S15_MS24*) and (iii) images with different Superior-to-Inferior coverage, e.g. including the brain (*S4_HC15*) or thoraco-lumbar levels (*S20_MS101*).

As reported in Table 1 (B.), the proposed spinal cord framework achieved significant superior results compared to *PropSeg*, with a median (IQR) Dice of 94.6 (4.6) versus 87.9 (18.3)% (p<0.001). In particular, the proposed method outperformed *PropSeg* in patients with severe cord atrophy in terms of (i) Dice: 92.9% versus 82.0% and (ii) relative volume difference: -3.6% versus +13.3%. The proposed framework was robust to MS-related pathology since the automatic segmentation yielded similar results between controls and MS subjects (median Dice: 95.2% versus 94.1%). The model generalized well to data from two sites unseen during the training (median Dice: 93.3%). For a typical $T_2$-w acquisition (matrix size: 384x384x52, resolution: 1mm isotropic), the computation time on an iMac (i7 4-cores 3.4 GHz 8Gb RAM), including reading and writing tasks, was 1min 55s for the proposed method versus 32s for *PropSeg*.

## 3.3. MS lesion segmentation

Figure 5 depicts several qualitative examples of MS lesion segmentations (both manual and automatic) from the testing data set. The main divergence between manual and automatic segmentations were located near normal-appearing structures (e.g. cerebrospinal fluid, grey matter) where the partial volume effect challenged tissue delineation (e.g. samples *S1_SPMS9, S2_RRMS5*). However, a visual inspection of the results shows that the network successfully learned the pattern of the normal-appearing grey matter despite its confounding intensities with MS lesions (e.g. samples *S7_RRMS14*). Instances where the automatic method correctly detected small lesions as well as lesions in atrophied cord are also shown in Figure 5 (see *S1_RRMS17, S2_CIS1, S8_PPMS10*). Although the Dice metric is widely used for medical image segmentation, it should be noted that it has a larger dynamic sensitivity to small versus large objects (see *S2_CIS1, S3_RRMS7).*

Table 1 (C.) shows the medians and IQRs of the metrics evaluating the automatic MS lesion segmentation. When pooling $T_2$-w and $T_2^*$-w, the automatic segmentation method reached a median (IQR) Dice of 60.0 (21.4)%. While this result might appear weak, it should be seen in light of the inter-rater study, where the raters achieved a median Dice against the "majority voting" masks of 60.7% compared to 56.8% for the automatic method. In terms of volumetric considerations, the automatic method provided satisfactory results, exhibiting a median relative volume difference of -14.5% (i.e. tends to under-segment the lesions). Median voxel-wise precision and sensitivity were 60.5% and 55.9%, respectively. Regarding the lesion-wise detectability, the automatic method yielded a low number of false positive (median precision: 76.9%) and false negative (median sensitivity: 83.3%) lesion labels per volume. The method was notably sensitive in detecting lesions on $T_2$-w sagittal scans (median sensitivity: 100.%). When confronted with data from sites excluded from the training data set, the method provided similar results as other sites (median sensitivity: 100.0%, median Dice: 57.0%). Finally, the automatic lesion detector yielded a volume-wise specificity of 88.6% on healthy control data, although 66.7% on MS data without any intramedullary lesions according to the raters.

Figure 6 compares the raters and automatic MS lesion segmentation on 10 testing subjects. An inter-rater variability was observed: the Dice results against the "majority voting" masks varied by 85.0% among the raters for subject 004, and by 21.0% for subject 008 (see Figure 6 A.). The disagreements between raters mainly occurred on the borders of the lesions, in particular, the lesion extension within the grey matter area on $T_2^*$-w images (see Figure 6 B.). The average time for manually segmenting lesions in one subject (two volumes per patient) was 18.7 minutes vs. 3.6 minutes using the automatic method (iMac i7 4-cores 3.4 GHz 8Gb RAM).

| A. Centerline Detection | Mean Square Error [mm] *Best value: 0* | | Localization Rate [0-100]% *Best value: 100* | |
|---|---|---|---|---|
| | *SVM+OptiC* | *$CNN_1$+OptiC* | *SVM+OptiC* | *$CNN_1$+OptiC* |
| **$T_1$-w Data** | **11.1 (11.8)** | **0.9 (0.5)** | **33.3 (48.9)** | **100 (0)** |
| **$T_2$-w Data** | **9.1 (12.8)** | **1.0 (0.9)** | 100. (33.3) | 99.7 (4.2) |
| **$T_2$*-w Data** | 0.9 (0.3) | 1.0 (0.6) | 100 (0) | 100 (0) |

| B. Spinal Cord Segmentation | Dice Coefficient [0-100]% *Best value: 100* | | Relative Volume Difference [0-100]% *Best value: 0* | |
|---|---|---|---|---|
| | *PropSeg* | *$CNN_{2\text{-}SC}$* | *PropSeg* | *$CNN_{2\text{-}SC}$* |
| **$T_1$-w Data** | **92.0 (13.5)** | **95.9 (1.5)** | -4.4 (11.1) | -0.3 (5.7) |
| **$T_2$-w Data** | **83.2 (18.6)** | **92.4 (5.1)** | 7.0 (26.8) | -0.2 (6.5) |
| **$T_2$*-w Data** | **94.1 (15.7)** | **95.5 (2.8)** | 4.3 (32.8) | -3.5 (9.8) |

| C. MS Lesion Segmentation | Dice Coefficient [0-100]% *Best value: 100* | Relative Volume Difference [0-100]% *Best value: 0* | Lesion-wise Sensitivity [0-100]% *Best value: 100* | Lesion-wise Precision [0-100]% *Best value: 100* | Voxel-wise Sensitivity [0-100]% *Best value: 100* | Voxel-wise Precision [0-100]% *Best value: 100* | Volume-wise Specificity [0-100]% *Best value: 100* |
|---|---|---|---|---|---|---|---|
| **$T_2$-w Data** | 57.6 (22.4) | -17.3 (61.3) | 90.0 (33.3) | 66.7 (58.3) | 51.4 (39.4) | 68.3 (39.6) | 80.6 |
| **$T_2$*-w Data** | 60.4 (25.0) | -4.5 (74.9) | 75.0 (47.2) | 100.0 (38.4) | 59.0 (38.6) | 47.4 (59.2) | 81.5 |

**Table 1:** Median (interquartile range) results, for the cord centerline detection (A), the spinal cord segmentation (B), and the MS intramedullary lesion segmentation (C). Results were computed from the testing data set, reported across contrasts. The best possible score value (i.e. not the best score reached) is indicated under each metric name. Performance comparisons between "*SVM+OptiC*" (Gros et al., 2018) and "*$CNN_1$+OptiC*", as well as between "*PropSeg*" (De Leener et al., 2015) and "*$CNN_{2\text{-}SC}$*" were statistically assessed using Kruskal-Wallis tests, and significant differences are indicated in bold ($p \leq 0.05$, adjusted with Bonferroni correction).

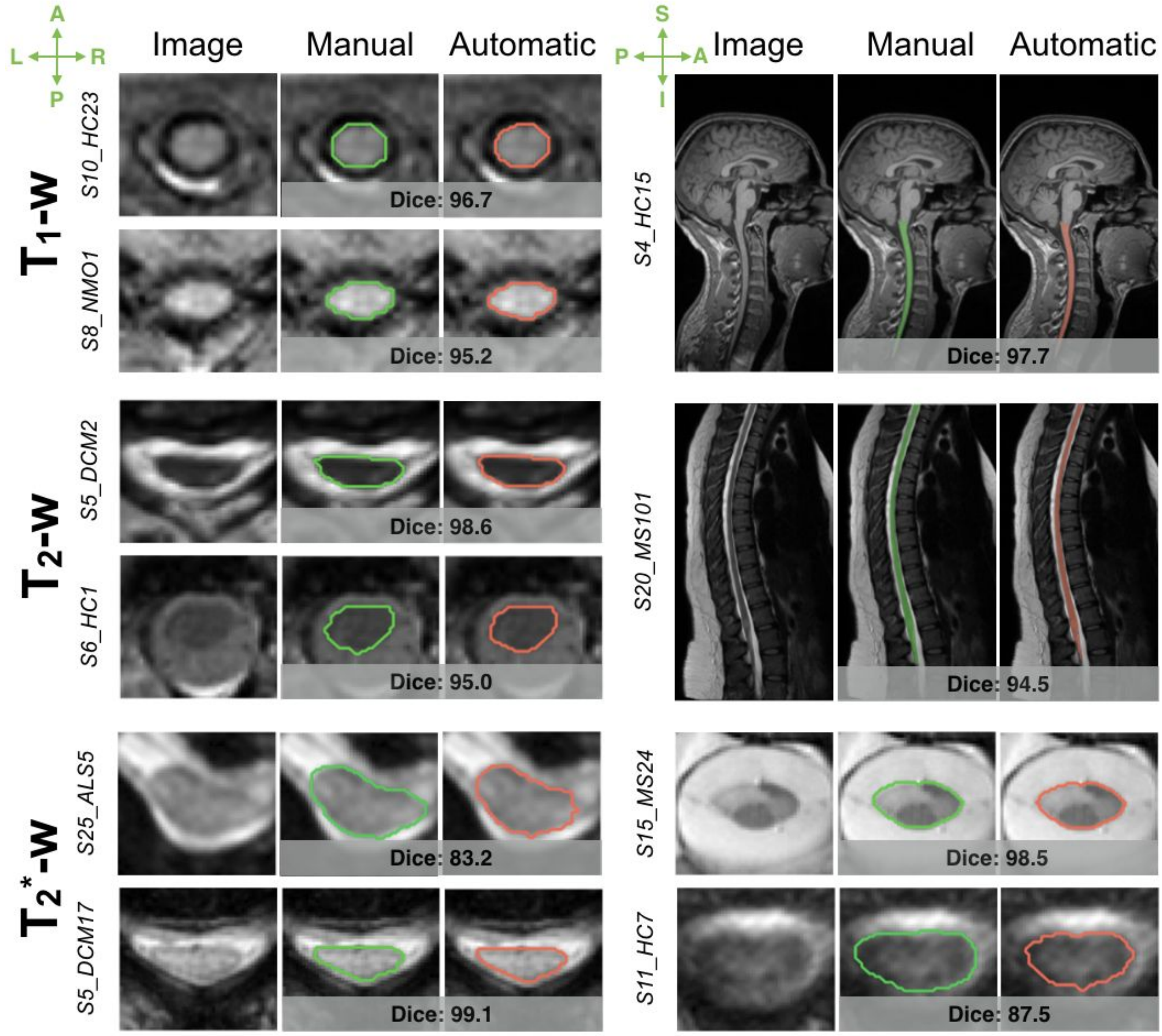

**Figure 4**: **Examples of automatic spinal cord segmentations** on $T_1$-w (top), $T_2$-w (middle) and $T_2^*$-w (bottom) MRI data. This includes a comparison between manual (green) and automatic (red) delineations, with Dice coefficient indicated just below each comparison. Note that the depicted samples represent a variety of subjects in terms of clinical status, and were scanned at different sites, identified by their ID (e.g. S10_HC23 is the ID of the HC subject #23, from the site #10). Abbreviations: A: Anterior ; P: Posterior ; L: Left ; R: Right ; I: Inferior ; S: Superior ; Auto.: Automatic ; HC: healthy controls ; MS: multiple sclerosis ; DCM: degenerative cervical myelopathy ; NMO: neuromyelitis optica ; ALS: amyotrophic lateral sclerosis.

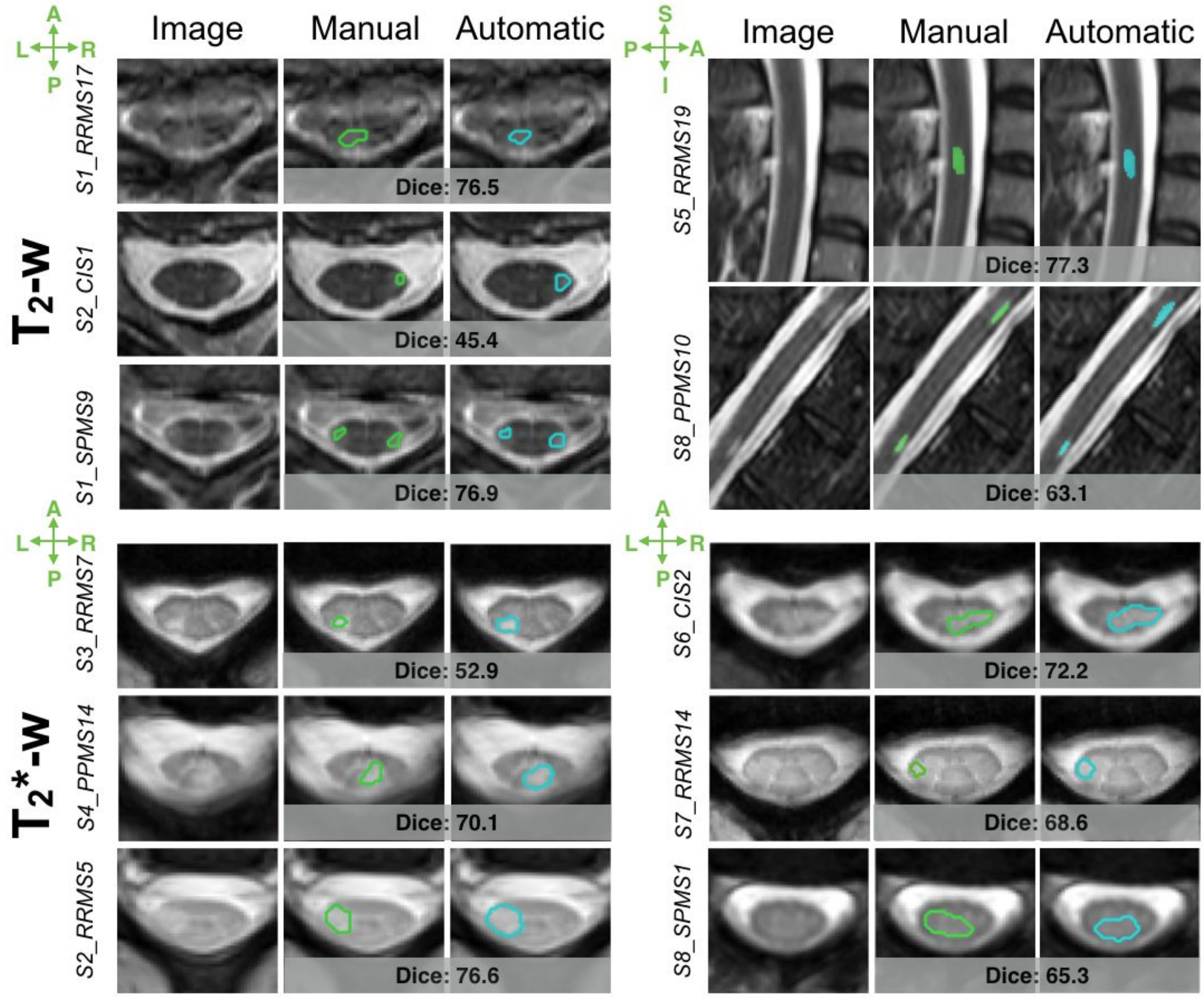

**Figure 5**: **Examples of automatic lesion segmentations** on Axial $T_2$-w (top left), Axial $T_2$*-w (bottom) and Sagittal T2-w (top, right) MRI data. This includes a comparison between manual (green) and automatic (blue) delineations, with Dice coefficients indicated just below each comparison. Note that the depicted samples were scanned at different sites, identified by their ID (e.g. S1_RRMS17 is the ID of subject #17 from site #1 with relapsing-remitting multiple sclerosis).

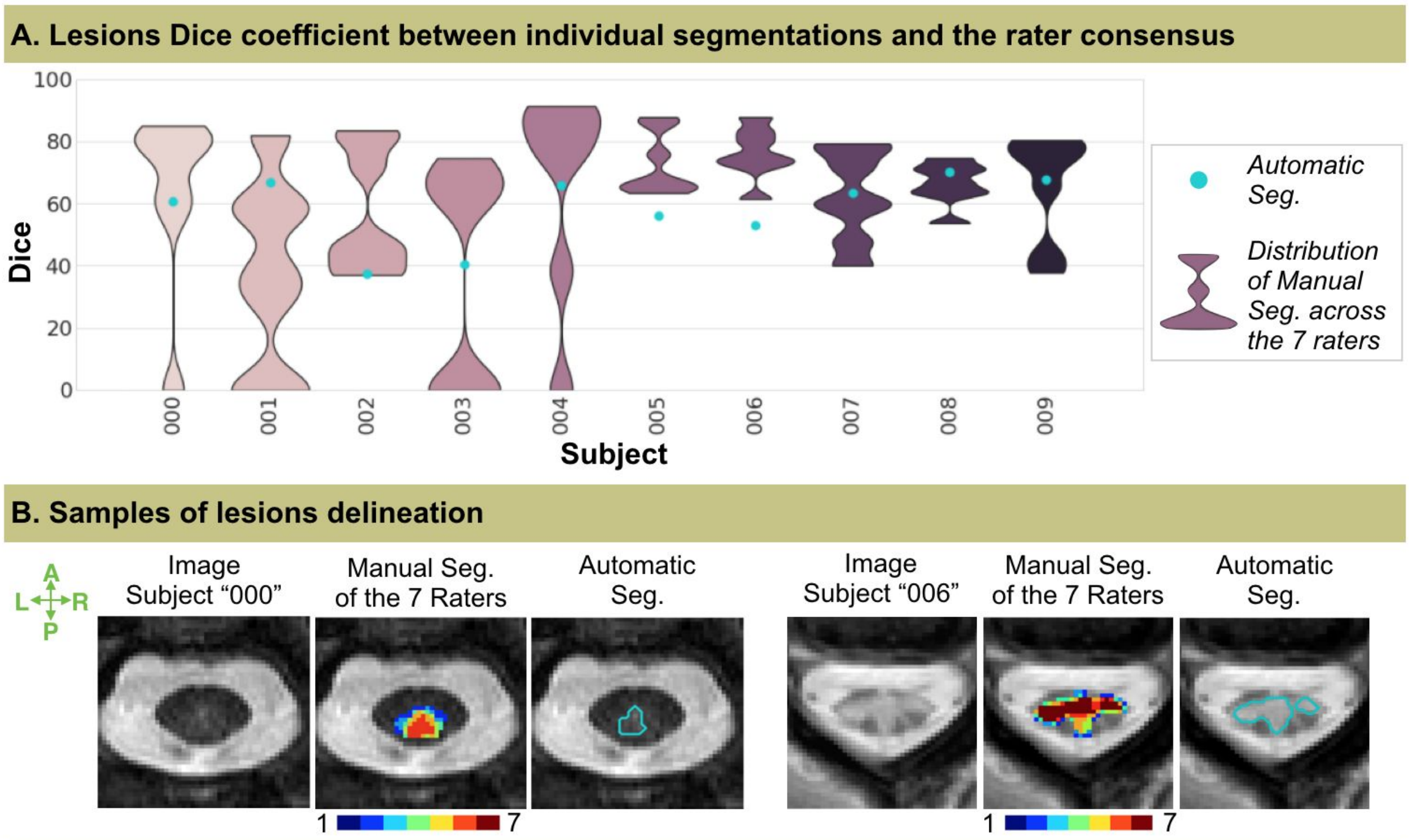

**C. Lesions segmentation time**

| Subject | Rater #1 | Rater #2 | Rater #3 | Rater #4 | Rater #5 | Rater #6 | Rater #7 | Auto. Seg. |
|---|---|---|---|---|---|---|---|---|
| Mean seg. time per subject [min] | Not Available | 17.5 | 24.5 | 11.2 | 6.5 | 15.5 | 36.8 | 3.6 |

**Figure 6**: **Inter-rater variability.** Comparison between raters and automatic MS lesion segmentation on 10 testing subjects. (A.) shows the Dice coefficient (range of [0-100] with 100% as best possible value) computed between the rater consensus (majority voting) and each individual rater (n=7) segmentation (purple distributions) as well as the automatic method (blue dot). (B.) depicts axial cross-sectional samples with the manual segmentation of the raters and the automatic delineation (blue). The consensus between raters vary from "low agreements" (in blues, mainly on the borders) to "strong agreement" (in reds, mainly on the cores). The green-to-red (see colormap) voxels were considered as part of the majority voting masks. (C.) presents the segmentation time, averaged across subjects, for each rater and the automatic segmentation (iMac i7 4-cores 3.4 GHz 8Gb RAM). Abbreviations: Seg.: Segmentation ; A: Anterior ; P: Posterior ; L: Left ; R: Right ; I: Inferior ; S: Superior ; Auto.: Automatic.

# 4. Discussion

We introduced a robust method to segment the spinal cord and/or intramedullary MS lesions. The proposed framework is based on a sequence of two CNNs, trained individually to tailor a set of specific filters for each target structure. The first network is trained to detect the spinal cord centerline within the 3D volume, so that the volume investigated by the second network is restricted to a close neighborhood of the target structures to segment (i.e. either the spinal cord or the intramedullary MS lesions). Furthermore, the framework has been designed to handle the heterogeneity of image acquisition features. Evaluation was performed on a large multi-site cohort including participants with various clinical conditions as well as healthy controls. The developed tools are freely available as part of SCT (De Leener et al., 2017a), version v3.2.2 and higher, through the functions sct_deepseg_sc and sct_deepseg_lesion.

## 4.1. Spinal cord centerline detection

Robustly localizing the spinal cord centerline on MRI data is a key step for automating spinal cord segmentation (De Leener et al., 2015; Horsfield et al., 2010) and template registration (De Leener et al., 2018; Stroman et al., 2008). The proposed method works in two steps: (i) recognition by a CNN of the spinal cord pattern on axial slices,(ii) regularisation of the spinal cord centerline continuity along the Superior-to-Inferior direction using a global curve optimisation algorithm (Gros et al., 2018). Although the spinal cord pattern was well identified by $CNN_1$ in the first step, resorting to the curve regularisation (step ii) was important for ensuring centerline consistency. This was especially true for patients with spinal cord atrophy, for whom the contrast between the cerebrospinal fluid and the spinal cord was frequently very low in large sections of the cord. Having produced detections of similar accuracy for axial and sagittal scans, this approach demonstrated its robustness to image resolution, especially when compared to its predecessor (Gros et al., 2018). In particular, $CNN_1$ enables a robust centerline detection on sagittal $T_2$-w images, which was often unsatisfactory with the SVM, likely due to the lack of variability in its training set ($n_{vol.}$=1) to apprehend the distortions of spinal cord shape when these images are resampled at $(0.5)^2mm^2$ in the cross-sectional plane. In addition, the new method can be used to separate spine and brain sections, which are regularly covered during cervical scans.

### 4.1.1. Limitations

The introduction of a detection step prior to the segmentation module was motivated by the high class imbalance (proportion of spinal cord and/or lesion compared to the rest of the volume) and the large heterogeneity of image features (contrast, field of view, etc.). However, the disadvantage of the sequential approach is that the segmentation framework is sensitive to the quality of the detection module. Fortunately though, the high performance of the spinal cord detection (median MSE of 1mm) is reliable enough to be cascaded by another CNN. When scans incorporated the brain, 13% of the spinal cord centerlines extended above the pontomedullary junction, but without impacting the consecutive cord segmentation.

#### 4.1.2. Perspectives

Besides the three MR contrasts investigated in this study ($T_1$-, $T_2$-, and $T_2^*$-w), we plan to cover other commonly-used sequences, such as diffusion-weighted scans and $T_2^*$-w echo-planar imaging (typically used for fMRI studies), and to make the additional trained models available in SCT. Apart from segmentation purposes, the centerline spatial information could guide an automatic tool for identification of the vertebral discs along the spinal canal (Ullmann et al., 2014), provide spinal cord curvature information for studying the biomechanics of the spine and planning surgery (Gervais et al., 2012; Little et al., 2016), or be used for localized shimming (Topfer et al., 2018, 2016; Vannesjo et al., 2017).

### 4.2. Spinal cord segmentation

Spinal cord segmentation has important clinical value for measuring cord atrophy in MS patients (Dupuy et al., 2016; Kearney et al., 2014; Losseff et al., 1996; Lundell et al., 2017; Rocca et al., 2013, 2011; Singhal et al., 2017). Besides MS pathology, spinal cord segmentation could provide a valuable quantitative assessment of spinal cord morphometry in the healthy population (Fradet et al., 2014; Papinutto et al., 2015) or be used as a biomarker for other spinal cord diseases (Martin et al., 2017; Nakamura et al., 2008; Paquin et al., 2018). We proposed an automatic method to segment the spinal cord, and validated the method against manual segmentation on a multi-site clinical data set involving a variety of pathologies. We also compared this method to the previously published *PropSeg* method (De Leener et al., 2015). The proposed method achieved better results than *PropSeg* in terms of Dice and relative volume difference, especially in patients with severe cord compression. When cerebrospinal fluid/spinal cord contrast is low (e.g. compressed cord), *PropSeg* tends to cause segmentation leakage, while CNN benefits from a larger spatial view (e.g. to detect vertebra edges) and performs better in those difficult cases. The segmentation performed well across 3 different MR contrasts ($T_1$-, $T_2$-, and $T_2^*$-w), without assuming a particular field of view, orientation or resolution (thanks to automatic preprocessing steps).

When presenting our model with data from new sites, performance was similar to when the data came from the original sites (i.e. sites included in the supervised learning). The ability of our model to generalise is likely due to the large training data set, mostly composed of 'real-world' clinical data and spanning a broad diversity of scanning platform and acquisition parameters (e.g. isotropic and anisotropic images, with both axial and sagittal orientations).

#### 4.2.1. Limitations

The requirement for a large training data set is both a blessing and a curse. While the large size and heterogeneity played a key role in the ability of the model to generalise, it also has a few downsides: (i) need time and expert knowledge for manually labeling a large amount of data, (ii) when the data are not available for sharing (due to ethical constraints), it prevents reproducibility, and (iii) the heterogeneity of the dataset hampers the performance when compared to when the model is trained and applied on an homogeneous dataset. To mitigate this issue, models trained here were made publicly available and can be fine-tuned with lesser amount of data (Ghafoorian et al., 2017a; Pan and Yang, 2010) for other specific applications (e.g., animal data, other pathologies, other MR contrasts).

Though the deformable model of *PropSeg* could be adjusted in cases of segmentation failure (e.g. alter the radius of the SC, or conditions of the deformation), there is less room with the CNN-based approach for changing input parameters during inference. Moreover, the presented method is slower than *PropSeg*, mainly due to the use of 3D convolutions (see section 4.4.2). It is, however, important to note that the evaluation was biased in favour of *PropSeg*, since most of the manual spinal cord delineations were produced by correcting the mask previously generated by *PropSeg*.

### 4.2.2. Perspectives

To improve image quality and reduce the variability across sites, preliminary experiments explored the impact of advanced preprocessing techniques, such as denoising (Coupe et al., 2008) and bias field correction (Tustison et al., 2010). Finding a set of generic preprocessing hyper-parameters that works for every data set is challenging. Preprocessing, fine-tuned for a specific and homogeneous data set, however, could improve the segmentation. Along with the spinal cord, the automatic segmentation of the cerebrospinal fluid could also provide a measure of the spinal canal volume for normalising cord volumes across people of different sizes, analogous to brain parenchymal fraction or brain to intra-cranial capacity ratio. Finally, the scan-rescan reproducibility of the proposed segmentation method will be the subject of future investigations.

## 4.3. MS lesion segmentation

Automating spinal cord MS lesion segmentation provides an efficient solution to evaluate large data sets for lesion burden analyses. A thorough search of the relevant literature did not yield available related work. Results of the automatic segmentation were similar to the inter-rater results, with the advantage of higher efficiency and reproducibility (i.e. the algorithm will always produce the same segmentation for the same image). While the Dice scores were relatively low (median: 60.0%), it should be noted that this metric is highly sensitive to the total lesion load and lesion sizes (Guizard et al., 2015; Harmouche et al., 2015; Styner et al., 2008). The median Dice of 60.7% between each rater and the consensus reading illustrates that point well, which is in line with recent inter-rater variability results obtained on brain lesions: 63% (Carass et al., 2017) and 66% (Egger et al., 2017). We also computed object-based metrics (i.e. lesion-wise precision and sensitivity) which are less subjective to lesion borders (Geremia et al., 2011; Harmouche et al., 2015; Lladó et al., 2012; Styner et al., 2008; Valverde et al., 2017a). In addition, monitoring the lesion count in the spinal cord is an important measure of disease activity, since each central nervous location where a new lesion appears would represent an entry point of the immune cells that mediate the inflammatory-demyelinating process, (i.e. a breach of the blood brain-barrier). In the clinical setting, intramedullary lesion count provides complementary information to what is obtained by brain lesion monitoring (Healy et al., 2017; Thompson et al., 2018). The relative volume difference was also reported since the total lesion volume is often used as a clinical biomarker.

### 4.3.1. Limitations

False positives and/or false negatives were likely due to the partial volume effect between the cord and cerebrospinal fluid, and mostly observed with small lesions (< $50mm^3$), which are also essential for MS disease staging, prognosis, and during clinical trials. Results of the automatic method, as well as the raters' assessments, hinted at variable levels of detectability across sites. Variations in sequences and image contrast are probably accountable for the observed differences in performance. We noticed an ability to generalise well to data exhibiting features which were absent in the training data, however the method is likely to perform best on data acquired with parameters similar to the training data (see Table A1). Recent initiatives to standardise spinal cord MRI acquisition (Alley et al., 2018), with spinal cord multi-parametric protocols available for the three main vendors (www.spinalcordmri.org/protocols), will likely help reducing such variability in the future.

Although the algorithm showed a good specificity overall when encountering lesion-free data, it is however important to note the difference in volume-wise specificity between healthy control data and MS data without intramedullary lesions. For healthy control data, low lesion volumes were segmented in the few false positive cases (median: $10.6mm^3$), which we observed to be largely induced by partial volume effects. Interestingly, the segmented lesion volumes were much larger in the false positive cases of the MS data (median: $150.5mm^3$), which is unlikely to be due to partial volume effects alone and could be owing to misdetections in the manual segmentations. Using data acquired with isotropic resolution (to minimise partial volume effect in one direction) and/or CNN architectures based on multimodal data (Havaei et al., 2016) would likely reduce the false positive rate and can be investigated in future studies (see also the next section below).

Lesion borders can often be diffuse, so that defining an "edge" can be somewhat arbitrary and highly subjective in these cases. As a result, lesion borders are frequently the site of disagreement between manual and automatic delineations, as well as among raters. This motivated our implementation of a data augmentation module to prompt the model to be less confident of the lesion border prediction (random and local erosion/dilation of the lesion masks during the training). Its specific effect on the segmentation performance will be validated in future work. Another promising avenue would be to include an uncertainty measure for lesion delineation (Nair et al., 2018), which could allow radiologists to refine lesions with high boundary-uncertainty.

### 4.3.2. Perspectives

In this work, MS lesion segmentation was achieved by processing each 3D scan independently, which is arguably a non-optimal use of the different available contrasts. In clinical settings however, it is not uncommon to have more than one acquisition covering the same region. Future work could consider recent advances in domain-adaptation (Ghafoorian et al., 2017b; Valindria et al., 2018) to overcome variations in imaging protocols. Indeed, a combination of the information from different MR contrasts should help the identification of very small lesions while reducing the number of false positives. The false positives could also be limited by extending the training data set with non-MS lesions (e.g. spinal cord injury), while generalising the lesion detector to other clinical conditions.

Considering that image labelling is time consuming and tedious, semi-supervised learning approaches should be explored to take advantage of the wide number of available unlabeled data (Baur et al., 2017). Another interesting avenue would be to explore patterns that have been automatically learned by the CNN (see Figure 7), as suggested by a recent study on brain lesions (Kamnitsas et al., 2017). For example, we were surprised by the ability of the network to distinguish lesions in the normal-appearing grey matter on $T_2$*-w scans, suggesting that the pattern of the healthy grey matter has been self-learnt. This observation could suggest that great potential lies in the combination of the CNN discriminative ability and clinical knowledge, such as spatial priors for cervical lesions (Eden et al., 2018). This is in line with previous segmentation work, where performance of traditional classifiers was significantly improved by incorporation of tissue priors (Harmouche et al., 2015; Shiee et al., 2010; Van Leemput et al., 1999). It would thus be interesting to investigate ways for encoding such available prior information into the network's feature space, so that clinical knowledge could direct the network towards the optimal solution. This could indeed drastically simplify the optimisation problem and mitigate false positive detections.

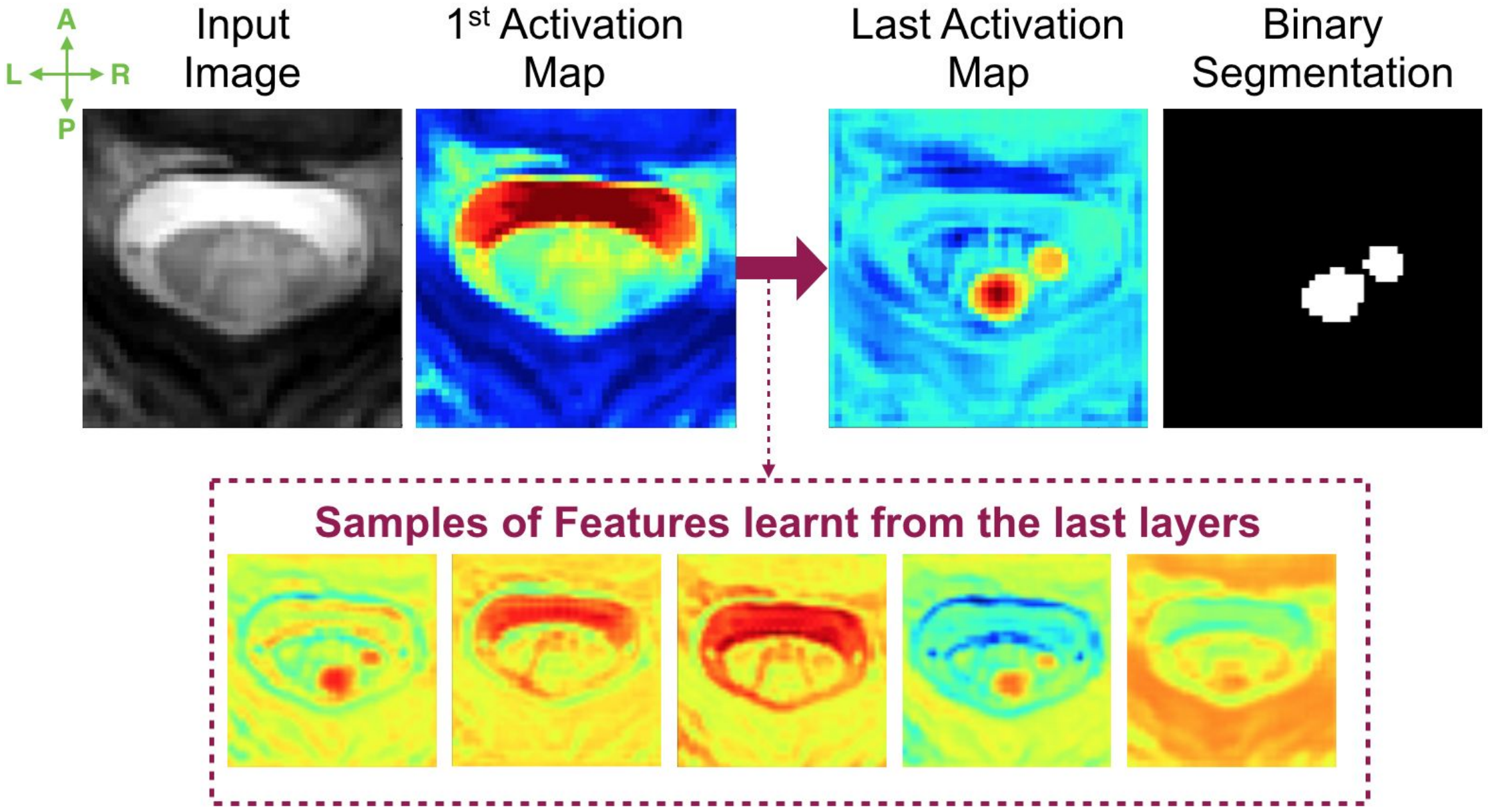

**Figure 7:** Visualisation of feature map instances, learnt by different layers of the $CNN_{2\text{-Lesion}}$, applied to an input image (left) leading to a binary segmentation (right). The normalised values represent the responses to filters learnt during the training step, with a colormap from blues (weak filter match) to reds (strong filter match).

## 4.4. CNNs Training

Due to the large heterogeneity in MRI contrast (see Figure 2), images were distributed among three MRI contrast data sets, for both the training and inference of the CNNs: (i) "$T_1$-weighted like" (i.e. dark cerebrospinal fluid / light cord), (ii) "$T_2$-weighted like" (i.e. light cerebrospinal fluid / dark cord / grey matter not visible), (iii) "$T_2^*$-weighted like" (i.e. light cerebrospinal fluid / dark cord / grey matter visible). The performance of the framework was consistent when trained with the 3 different MR contrast data sets, which highlights its robustness to different training conditions.

### 4.4.1. Class imbalance

An important challenge to the design of automated MS lesion segmentation methods is the extremely unbalanced nature of the data. In this work, this issue of class imbalance was mitigated by using the Dice loss, by performing an extensive data augmentation, and by restricting the search around the spinal cord centerline thanks to $CNN_1$.

In preliminary experiments, we explored the benefit of under-sampling the negative class during the training to address the massive class imbalance. While it significantly facilitated the training convergence, it biased the classifier towards the positive class and may have resulted in a drastic increase in false positive detections. More complex sampling schemes (Havaei et al., 2015; Jesson et al., 2017; Valverde et al., 2017a), successfully employed in medical image segmentation or detection tasks, could be investigated for spinal cord applications.

Moreover, in exploratory experiments, we also tested various loss functions specifically proposed to mitigate the class imbalance issues: the weighted cross-entropy (Ronneberger et al., 2015), the Dice (Milletari et al., 2016), and the "sensitivity - specificity" (Brosch et al., 2015) loss functions. Although the Dice loss caused narrow boundaries of confidence intervals at the edge, it yielded better results. In the future, other loss functions, fashioned to handle highly unbalanced data sets, could be tested, such as the Focal Loss (Lin et al., 2017) or the Generalised Dice overlap (Sudre et al., 2017).

### 4.4.2. 3D spatial information

Prior experiments also explored the use of 3D instead of 2D patches, as they were preferred in recent work on biomedical volumes (Çiçek et al., 2016; Kamnitsas et al., 2017; Milletari et al., 2016). However, while 3D patches provide more context-rich information, 3D CNNs have more parameters, and thus more memory and computational constraints.

For the spinal cord detection step, 2D patches were used to localize the position of the cord. Two-dimensional axial patches were adopted here for the sake of computational simplicity, considering that 3D patches did not yield substantial improvements. The use of 2D dilated convolutions might account for the accurate detections. Indeed, by increasing the receptive fields, dilated convolutions benefit from a broader spatial context for detecting sparse structures, while maintaining a relatively low number of parameters to optimise.

In most cases, the spinal cord segmentation quantitative results were similar whether 2D or 3D patches were used. However, in the cases with exceptional lesion load and severe atrophy, the incorporation of 3D contextual information showed noteworthy improvements, which consequently motivated the adoption of 3D patches. As mentioned before, the use of 3D convolutions caused a drastic increase of memory consumption, computational cost and training time. Further studies could investigate solutions to reduce the memory consumption, such as the Reversible Residual Network architecture (Gomez et al., 2017) or multi-stream architectures (Prasoon et al., 2013). Furthermore, future work could explore the benefit of 3D dense conditional random fields (Christ et al., 2016; Krähenbühl and Koltun, 2011; Zheng et al., 2015) to incorporate 3D context instead of using 3D convolutions.

# 5. Conclusion

We presented an original automated spinal cord and MS lesion segmentation method, based on a sequence of two convolutional neural networks. Spinal cord segmentation results outperformed a state-of-the-art method on a multi-site and highly heterogeneous clinical data set. Lesion segmentation results were generally within the range of manual segmentations, although the false positive rate warrants further investigations. The presented automatic methods are open-source and readily accessible in SCT (version v3.2.2 and higher).

## Acknowledgements:

The following people are acknowledged for MRI acquisition: Manuel Taso, Jamie Near, Ives Levesque, Guillaume Gilbert, Robert Barry, Johanna Vannesjo, Antonys Melek, and Charles Tremblay. The following people are acknowledged for sharing data: Eric Klawiter (Massachusetts General Hospital), Julius Dewald, Haleh Karbasforoushan (Northwestern University), Pierre-François Pradat and Habib Benali (Pitié-Salpêtrière Hospital), Barry Bedell (Biospective), Claudia AM Gandini Wheeler-Kingshott (University College London), Pierre Rainville (Université de Montréal), Bailey Lyttle, Benjamin Conrad, Bennett Landman (Vanderbilt University), Maryam Seif and Patrick Freund (Spinal Cord Injury Center Balgrist, University Hospital Zurich), Seok Woo Kim, Jisun Song, Tom Lillicrap, and Emil Ljungberg.

We acknowledge the NVIDIA Corporation for the donation of a GPU.

We would like to warmly thank the members of NeuroPoly Lab for fruitful discussions and valuable suggestions, especially Harris Nami and Ryan Topfer for reviewing the manuscript, and Christian Perone and Francisco Perdigón Romero for their inputs on deep learning.

## Grant Support:

Funded by the Canada Research Chair in Quantitative Magnetic Resonance Imaging (JCA), the Canadian Institute of Health Research [CIHR FDN-143263], the Canada Foundation for Innovation [32454, 34824], the Fonds de Recherche du Québec - Santé [28826], the Fonds de Recherche du Québec - Nature et Technologies [2015-PR-182754], the Natural Sciences and Engineering Research Council of Canada [435897-2013], IVADO, TransMedTech and the Quebec BioImaging Network, ISRT, Wings for Life (INSPIRED project), the SensoriMotor Rehabilitation Research Team (SMRRT), the National Multiple Sclerosis Society NMSS RG-1501-02840 (SAS), NIH/NINDS R21 NS087465-01 (SAS), NIH/NEI R01 EY023240 (SAS), DoD W81XWH-13-0073 (SAS), the Intramural Research Program of NIH/NINDS (JL, DSR, GN), the Centre National de la Recherche Scientifique (CNRS), The French Hospital Programme of Clinical Research (PHRC) for the EMISEP project, ClinicalTrials.gov Identifier: NCT02117375, the “Fondation A*midex-Investissements d'Avenir” and the “Fondation Aix-Marseille Université”, the Stockholm County Council (ALF grant 20150166), a postdoc fellowship from the Swedish Society for Medical Research (TG), a postdoc non-clinical fellowship from Guarantors of Brain (FP), the French State and handled by the "Agence Nationale de la Recherche", within the framework of the "Investments for the Future" programme, under the reference ANR-10-COHO-002 Observatoire Français de la Sclérose en plaques (OFSEP), with the assistance of Eugène Devic EDMUS Foundation against multiple sclerosis; EDMUS, a European database for multiple sclerosis. Confavreux C, Compston DAS, Hommes OR, McDonald WI, Thompson AJ. J Neurol Neurosurg Psychiatry 1992; 55: 671-676, NIH/NINDS R21 NS087465-01 (SAS), NIH/NEI R01 EY023240 (SAS), DoD W81XWH-13-0073 (SAS), Grant MOP-13034, National Multiple Sclerosis Society NMSS RG-1501-02840 (SAS). Additional funding sources include NIH/NINDS R21 NS087465-01 (SAS), NIH/NEI R01 EY023240 (SAS) and DoD W81XWH-13-0073 (SAS).

# Appendix

**Table A1.** Summary of MRI systems, acquisition parameters, and vertebral coverage across sites contributing more than 20 subjects to this study.

| Site | MRI scanner | Contrast, Orientation | Vertebral coverage (median range) | TR (ms) | TE (ms) | FOV ($mm^2$) | Number of slices, slice thickness (mm) |
|---|---|---|---|---|---|---|---|
| Aix-Marseille University, Hôpital La Timone, Marseille, France (n = 61) | Siemens Verio 3T | $T^*_2w$, Axial | C1-C7 | 849 | 23 | 179x179 | 40, 3.00 |
| | | $T_2w$, Sagittal | C1-C7 | 3000 | 68 | 261x261 | 15, 2.50 |
| | | 3D $T_1w$ | C1-L5 | 2260 | 2.09 | 384x264 | 176, 1.00 |
| | | 3D $T_2w$ | C1-L5 | 1500 | 119 | 257x186 | 51, 1.00 |
| Brigham and Women's Hospital, Boston, USA (n = 84) | 3 T | $T_2w$, Axial | C1-C7 | 5070 | 101 | 179x179 | 47, 3.00 |
| Karolinska University Hospital, Stockholm, Sweden (n = 53) | Siemens Trio 3T | $T_2^*w$, Axial | C1-C7 | 561 | 17 | 179x179 | 30, 4.40 |
| Massachusetts General Hospital, Boston, USA (n = 38) | 7 T | $T_2^*w$, Axial | C1-C7 | 500 | 7.8 | 219x210 | 36, 3.00 |
| National Institutes of Health Clinical Center, Maryland, USA (n = 35) | Siemens Skyra 3T | $T_2^*w$, Axial | C1-C7 | 560 | 17 | 260x195 | 28, 5.00 |
| | | $T_2w$, Sagittal | C1-C7 | 6000 | 27 | 384x384 | 30, 1.00 |
| NYU Langone Medical Center, New York, USA (n=30) | 3T | $T_2w$, Axial | C1-T3 | NA | NA | 200x156 | 60, 4.86 |
| | | $T_2w$, Sagittal | C1-T4 | NA | NA | 180x135 | 32, 3.90 |
| Pitié-Salpêtrière Hospital, France (n=70) | Siemens Trio 3T | $T_2^*w$, Axial | C1-C6 | 470 | 17 | 180x180 | 23, 3.00 |
| | | 3D $T_2w$ | C1-T3 | 1500 | 120 | 280x280 | 52, 0.90 |
| French Observatory of Multiple Sclerosis, France (n = 59) | 3 T | $T_2^*$w, Axial | C1-C3 | 992 | 29 | 198x179 | 16, 4.55 |
| | | $T_2$w, Sagittal | C1-C7 | 4720 | 74 | 338x338 | 12, 4.80 |
| San Raffaele Scientific Institute, Vita-Salute San Raffaele University, Milan, Italy (n = 118) | Philips Achieva 3T | $T_2^*$w, Axial | C1-C7 | 47 | 6.5 | 150x150 | 40, 2.50 |
| | | $T_2$w, Sagittal | C1-C7 | 2933 | 70 | 250x250 | 14, 2.50 |
| Toronto Western Hospital, Canada (n=88) | GE Healthcare Signa Excite 3T | $T_2^*$w, Axial | C1-C7 | 650 | 5, 10, 15 | 200x200 | 12, 4.00 |
| | | 3D $T_2$w | C1-T4 | 5400 | 2600 | 200x200 | 62, 0.80 |
| University Hospital of Rennes, Rennes, France (n = 71) | Siemens Verio VB17 3T | $T_2^*$w, Axial | C1-C7 | 849 | 23 | 179x179 | 40, 3.30 |
| | | $T_2$w, Sagittal | C1-C7 | 3000 | 68 | 261x261 | 15, 2.75 |
| University College London, London, UK (n = 50) | 3 T | $T_2^*$w, Axial | C1-C3 | 23 | 5 | 240x240 | 10, 5.00 |
| | | $T_2$w, Sagittal | C1-C7 | 4000 | 80 | 256x256 | 12, 3.00 |
| Functional Neuroimaging Unit (UNF), Montreal,, Canada (n=113) | Siemens Trio 3T | 3D $T_2$w | C1-L5 | 1500 | 119 | 385x160 | 51, 1.00 |
| | | $T_2^*$w, Axial | C2-C5 | 539 | 5.41, 12.56, and 19.16 | 160x160 | 10, 5.00 |
| | | $T_2^*$w, Axial | C4-C8 | 3050, 3200, and 3140 | 33 | 132x132 | 10, 9.00 |
| | | 3D $T_1$w | C1-L5 | 2260 | 2.09 | 320x240 | 192, 1.00 |
| Zuckerberg San Francisco General Hospital, San Francisco, USA (n = 26) | 3 T | $T_2^*$w, Axial | C1-C7 | 3516 | 72 | 179x179 | 36, 3.30 |
| Vanderbilt University Medical Center, Nashville, USA (n | Philips Achieva 3 T | $T_2^*$w, Axial | C2-C5 | 753 | 7 | 162x162 | 14, 5.00 |
| = 44) | | $T_2$w, Sagittal | C1-C7 | 2500 | 100 | 251x251 | 18, 2.00 |
| Xuanwu Hospital, China (n=53) | Siemens Trio 3T | 3D $T_1$w | C1-T6 | 1000 | 3 | 320x260 | 96, 1.00 |
| University Hospital Zurich, Switzerland (n=21) | Siemens Skyra 3T | $T_2^*$-w, Axial | C1-C4 | 44 | 19 | 192x162 | 20, 2.50 |

## Declaration of interest:

Charley Gros, Benjamin De Leener, Josefina Maranzano, Dominique Eden, Atef Badji, Govind Nair, Tobias Granberg, Hugh Kearney, Ferran Prados, Russell Ouellette, Daniel S. Reich, Pierre Labauge, Leszek Stawiarz, Anne Kerbrat, Elise Bannier, Shahamat Tauhid and Julien Cohen-Adad have no relevant financial interests to disclose.

Prof. Filippi is Editor-in-Chief of the Journal of Neurology; received compensation for consulting services and/or speaking activities from Biogen Idec, Merck-Serono, Novartis, Teva Pharmaceutical Industries; and receives research support from Biogen Idec, Merck-Serono, Novartis, Teva Pharmaceutical Industries, Roche, Italian Ministry of Health, Fondazione Italiana Sclerosi Multipla, and ARiSLA (Fondazione Italiana di Ricerca per la SLA).

Jan Hillert has received honoraria for serving on advisory boards for Biogen, Sanofi-Genzyme and Novartis; and speaker's fees from Biogen, Novartis, Merck-Serono, Bayer-Schering, Teva and Sanofi-Genzyme.; and has served as P.I. for projects or received unrestricted research support from Biogen Idec, Merck-Serono, TEVA, Sanofi-Genzyme and Bayer-Schering.

M.A. Rocca received speaker honoraria from Biogen Idec, Novartis, Genzyme, Sanofi-Aventis, Teva and Merck Serono and receives research support from the Italian Ministry of Health and Fondazione Italiana Sclerosi Multipla.

Dr. S. Narayanan reports personal fees from NeuroRx Research, a speaker's honorarium from Novartis Canada, and grants from the Canadian Institutes of Health Research, unrelated to the submitted work.

P. Valsasina received speaker honoraria from Biogen Idec, Novartis and ExceMED.

Donald G. McLaren and Vincent Auclair are currently employees of Biospective, Inc.

Rohit Bakshi has received consulting fees from Bayer, EMD Serono, Genentech, Guerbet, Sanofi-Genzyme, and Shire and research support from EMD Serono and Sanofi-Genzyme.

Jean Pelletier received speaker honoraria from Biogen, Roche, Genzyme, Novartis, and research supports from the French Ministry of Health and ARSEP.